%% file: main.tex
\pdfoutput=1
\documentclass{article}

\usepackage{iclr2027_conference,times}
\usepackage{amsmath,amssymb}
\usepackage{booktabs}
\usepackage{graphicx}
\usepackage{wrapfig}
\usepackage{float}
\usepackage{xcolor}
\usepackage{hyperref}
\usepackage{url}

\title{Reach or Solve? Deep Diving into Agentic RL Gains with Checkpoint Handoffs\thanks{Code: \url{https://github.com/Xuanxuana1/AgenticRL_Handoff}}}

\author{%
Xuan Liu\\
Shanghai Jiao Tong University\\
\texttt{liuxuan\_cn@outlook.com}
\And
Jingbin Qian\\
Rice University\\
\texttt{jingbinqian2002@gmail.com}}

\iclrfinalcopy

\begin{document}

\maketitle
\lhead{Preprint}

\begin{abstract}
Reinforcement learning (RL) is widely used to improve language-model agents,
and its gains are usually measured by final task success. However, an agent's
earlier actions shape the states in which its later decisions are made, so
final task success conflates the ability to reach useful states with the
ability to complete the task once there. Comparing agents only on the states
each one reaches does not separate the two, since each agent is then scored on
states selected by its own actions. To address this
conflation, we introduce checkpoint handoff, an
evaluation protocol that decouples reaching from completing without
retraining. One checkpoint acts as a \emph{reacher} up to a handoff point, and
another continues as the \emph{solver} from the same replayed history. In
detail, (1) \textsc{Reach} measures how often a reacher arrives at states that
a replayable environment verifies to be a fixed number of actions from
success, and (2) \textsc{Solve} measures how often a solver completes the task
from identical cloned copies of those states. Crossing supervised fine-tuning
(SFT) and RL checkpoints in both roles across two benchmarks and two
independently released training pipelines, we find that the gain from
switching the solver from SFT to RL is consistently larger when RL is the
reacher, at all three model scales on TravelPlanner and on both ALFWorld
splits. Further analyses on ALFWorld show that RL improves both \textsc{Reach}
and \textsc{Solve}. The solver gain is larger under an RL reacher because RL
reaches solvable states more often, and separately measured
\textsc{Reach} and \textsc{Solve} gaps recover most of this difference.
Because handoff only
requires replaying one checkpoint's history under another, agentic RL
evaluations can report arrival and completion alongside final success.
\end{abstract}

\section{Introduction}
\label{sec:introduction}

Language-model agents have rapidly become a central application of foundation
models. They search, call tools, and interact with environments over multiple
turns~\citep{xie2024travelplanner,yao2024taubench,shridhar2021alfworld}.
As these agents take on longer and more interactive tasks, the methods used to
train them have evolved accordingly~\citep{chu2025sft,zhang2026the}.
Specifically, supervised fine-tuning
(SFT) first develops these abilities by imitating demonstrations.
AgentTuning mixes agent
interaction trajectories with general instruction data~\citep{zeng2023agenttuning},
while Agent-FLAN redesigns the training corpus to separate format following
from agent reasoning~\citep{chen2024agentflan}. More recently, agentic
reinforcement learning (RL) improves a policy through interaction and reward, with
work studying trajectory-level optimization and stability
(RAGEN), transition-level credit assignment (Agent Lightning), progressively
longer interaction horizons (AgentGym-RL), and reusable skill libraries
(SkillRL)~\citep{wang2025ragen,luo2025agentlightning,xi2025agentgymrl,xia2026skillrl}.
Yet these advances leave unclear whether RL's gains over SFT come from
reaching more useful states, acting better from a given state, or both.
This distinction matters for understanding what RL learns and deciding which
part of training or evaluation to improve.

\begin{figure}[t]
  \centering
  \includegraphics[width=0.90\linewidth]{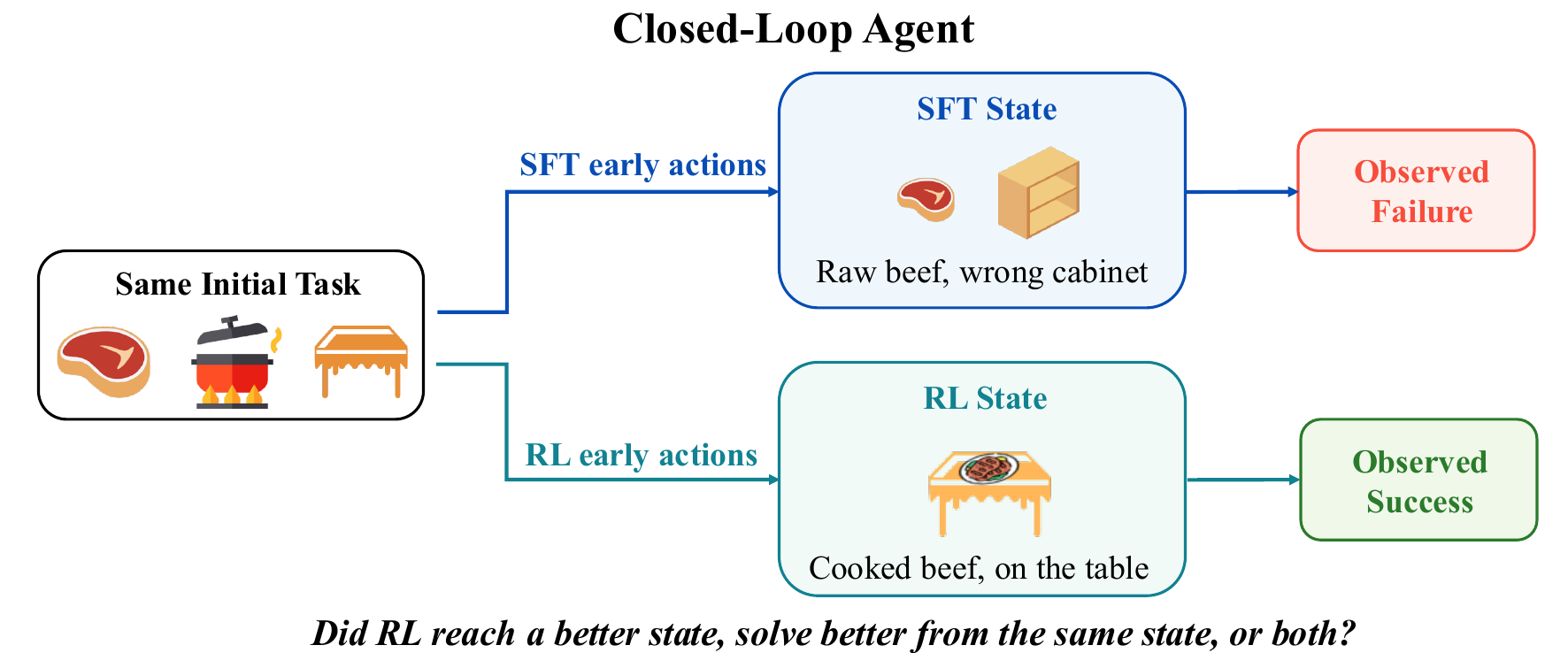}
  \caption{\textbf{The endogenous state problem.} Two checkpoints receive one
  task and act. Their early actions leave them in different states, and each
  is then scored from the state it produced.}
  
  \label{fig:endogenous-state}
\end{figure}

The difficulty arises because an agent's later inputs depend on its earlier
actions. In a closed loop, actions change the environment and hence subsequent
observations~\citep{sutton2018reinforcement}. SFT and RL checkpoints can thus
receive the same task while generating different histories and being scored
from different states. In Figure~\ref{fig:endogenous-state}, one checkpoint
reaches the wrong cabinet with raw beef, while the other has already cooked
and placed it. We call this policy dependence of evaluation inputs the
\emph{endogenous state problem}. Endpoint comparisons change both the states
reached and the actions taken from them, leaving the source of a gain
unresolved.

Several evaluation families, reviewed more fully in
Appendix~\ref{app:related}, look inside an endpoint. Progress evaluators score
subgoal completion, process evaluators assess intermediate decisions and
procedural compliance, and proxy-state methods verify the environment state
left by a tool call~\citep{ma2024agentboard,gritta2026process,chuang2026proxy}.
Exploration analyses measure entry into useful regions, while failure
diagnosis separates errors made before and after a capability becomes
exercisable~\citep{ye2026look,ji2026actguide,shao2026beyond}.
Causal decompositions further attribute outcomes to actions, transitions, or
perception~\citep{triantafyllou2025counterfactual,wang2026badseeing}.
\textbf{These methods provide useful descriptions of a checkpoint's realized rollout,
but they do not reveal how another checkpoint would continue from the same
state and history.} Comparing policies only on the states they reach does not
fix this. Arrival depends on both the policy and task difficulty, so such
filtering mixes solver ability with the difficulty of the surviving
tasks~\citep{hernan2004structural}. Restricting to tasks that both policies
reach narrows the target to a selected subset, and even then the two policies
need not arrive with the same history or budget.
Attributing the overall gain therefore requires comparing solvers from
identical states while retaining non-arrivals in the population-level
comparison.

To address this attribution gap, we propose \emph{checkpoint handoff}, which
intervenes on who continues from a state. Specifically, we introduce two
roles. A \emph{reacher} acts first and determines which states the agent
arrives at, and a \emph{solver} then continues from a state held fixed and
determines whether the task is finished. We cross
these roles over $\{\mathrm{SFT},\mathrm{RL}\}$ without retraining.
At an environment-verified frontier, we clone the environment, history, and
remaining budget and let both solvers continue from identical inputs.
\textsc{Reach} measures arrival frequency, while \textsc{Solve} measures
completion conditional on arrival. Non-arrivals remain failures in the
full-population endpoint comparison. This intervention separates
policy-induced state visitation from state-conditional
value~\citep{kakade2002approximately} on released checkpoints. Profiles
measured on disjoint data then test whether the two components account for
how much larger the solver gain is under an RL reacher.

Experimentally, we systematically study SFT/RL checkpoint pairs released by
two independent training pipelines on TravelPlanner and ALFWorld. We organize the comparison around three questions.
\textbf{RQ1} asks whether the gain from switching the solver from SFT to RL
depends on whether the reacher is SFT or RL. \textbf{RQ2} asks whether RL's
advantage over SFT comes from reaching the solvable frontier more often,
solving better from identical states, or both. \textbf{RQ3} asks whether
independently measured SFT/RL differences in \textsc{Reach} and \textsc{Solve}
account for the size of the RQ1 interaction.

This paper makes three contributions.
\begin{itemize}
  \item We characterize the endogenous state problem of closed-loop agents by
  decomposing endpoint success into state visitation and state-conditional
  value, and show that neither endpoint comparison nor filtering to reached
  states can attribute an RL gain.
  \item We propose checkpoint handoff, which separates \textsc{Reach} from
  \textsc{Solve} on released checkpoints without retraining, using an
  environment-verified frontier, cloned-state continuation, and an
  out-of-sample test that recombines independent \textsc{Reach} and
  \textsc{Solve} gaps.
  \item Experiments on two benchmarks and two pipelines show a positive
  reacher--solver interaction in every condition, and ALFWorld replay finds
  RL gains in both \textsc{Reach} and \textsc{Solve}.
\end{itemize}

\section{The Endogenous State Problem}
\label{sec:problem}

\subsection{Setting and Estimand}
\label{sec:setting}

Let a task $T$ be drawn from a probability distribution $p_{\mathcal T}$. At
step $t$, let $x_t$ be the environment configuration, $o_t$ the observation,
$a_t$ the action, $h_t=(o_0,a_0,\ldots,a_{t-1},o_t)$ the visible history, and
$b_t$ the remaining action budget. We call
\begin{equation}
  \Sigma_t=(T,x_t,h_t,b_t)\in\mathcal S
  \label{eq:evaluation-state}
\end{equation}
the evaluation state: the input from which the rest of the episode is scored.
The checkpoint sees only $h_t$; a replay-capable evaluator also retains $x_t$.
The episode ends at a step $t_{\mathrm{term}}$ by submission, termination, or
budget exhaustion, and a task verifier $g_T:\mathcal S\to\{0,1\}$ scores the
endpoint $Y=g_T(\Sigma_{\mathrm{term}})$, where
$\Sigma_{\mathrm{term}}=\Sigma_{t_{\mathrm{term}}}$. Let
$\mathcal C=\{\mathrm{SFT},\mathrm{RL}\}$ index released checkpoints and
$\pi_c(a\mid h)$, $c\in\mathcal C$, their induced policies on any visible
history $h$. Decoding is fixed, so we intervene on $c$ and read the
consequences off $\pi_c$. For a fixed state $\sigma\in\mathcal S$ and a random continuation
seed, let $Y(R;\sigma)\in\{0,1\}$ be the potential endpoint when checkpoint
$R\in\mathcal C$ continues from $\sigma$ to termination, in the sense of
potential outcomes~\citep{rubin1974estimating}; its mean
$V_R(\sigma)=\Pr(Y(R;\sigma)=1)$ is the continuation value, a property of the
state and the solver alone. For a fixed handoff distribution, solver
performance is its mean continuation value.

We define a handoff rule as a prespecified map from the reacher's trajectory
to a cut $\tau$, selected independently of solver assignment and continuation
outcomes. Running $W\in\mathcal C$ alone up to $\tau$
induces a distribution $d_W^{\tau}$ over $\Sigma_\tau$, and we let a second
checkpoint $R$ continue. The reacher $W$ sets the measure, the solver $R$ sets
the integrand,
\begin{equation}
  J_\tau(W,R)=\mathbb E_{\Sigma\sim d_W^{\tau}}\!\left[V_R(\Sigma)\right].
  \label{eq:reach-solve}
\end{equation}
Assignments with $W=R$ provide same-checkpoint controls; those with $W\ne R$
switch checkpoints at handoff.

\paragraph{The attribution question.}
The diagonal controls run one checkpoint in both roles and report
$J_\tau(c,c)$. Writing $J$ for $J_\tau$, the RL gain then splits
into a reach term and a solve term along either of two paths, by the same
telescoping that underlies the performance difference
lemma~\citep{kakade2002approximately}:
\begingroup\small
\begin{align}
  J(\mathrm{RL},\mathrm{RL})-J(\mathrm{SFT},\mathrm{SFT})
  ={}&\underbrace{J(\mathrm{RL},\mathrm{RL})-J(\mathrm{SFT},\mathrm{RL})}_{\text{reach, under the RL solver}}
  +\underbrace{J(\mathrm{SFT},\mathrm{RL})-J(\mathrm{SFT},\mathrm{SFT})}_{\text{solve, on SFT states}}
  \notag\\
  ={}&\underbrace{J(\mathrm{RL},\mathrm{SFT})-J(\mathrm{SFT},\mathrm{SFT})}_{\text{reach, under the SFT solver}}
  +\underbrace{J(\mathrm{RL},\mathrm{RL})-J(\mathrm{RL},\mathrm{SFT})}_{\text{solve, on RL states}}.
  \label{eq:attribution-paths}
\end{align}
\endgroup
Every right-hand term is a cross assignment in which one checkpoint continues
from states produced by the other. Endpoint evaluation observes only
$J_\tau(c,c)$, so it leaves $d_W^\tau$ and $V_R$ confounded; the two paths can
therefore disagree. The three research questions follow directly:
RQ1 varies the reacher while measuring the solver gain, RQ2 attributes RL's
advantage to reach and solve, and RQ3 tests the interaction prediction below.

\subsection{Endpoint Comparisons Change Two Factors at Once}
\label{sec:two-factors}

The same-checkpoint contrast on the left of Equation~\ref{eq:attribution-paths}
changes the measure and integrand together, so it cannot attribute the
difference to either. Evaluating all four assignments instead yields the
risk-difference interaction of a two-by-two factorial design~\citep{vanderweele2015explanation},
\begin{align}
  I_\tau={}&\big[J_\tau(\mathrm{RL},\mathrm{RL})-J_\tau(\mathrm{RL},\mathrm{SFT})\big]
  \notag\\
  &-\big[J_\tau(\mathrm{SFT},\mathrm{RL})-J_\tau(\mathrm{SFT},\mathrm{SFT})\big].
  \label{eq:factorial-interaction}
\end{align}
$I_\tau$ is the difference between the two solve terms of
Equation~\ref{eq:attribution-paths}, and equally between the two reach terms.
A positive $I_\tau$ means that switching the solver to RL gains more when RL
is also the reacher, so the two attribution paths disagree. It does not say
whether this comes from how often each reacher delivers a solvable state or
from how the solver gap varies across states
(Section~\ref{sec:cloned-state-cross}).

\subsection{Arrival Diverges Long Before Success Does}
\label{sec:pilot}

\begin{wrapfigure}{r}{2.4in}
  \vspace{-0.35cm}
  \centering
  \includegraphics[width=2.3in]{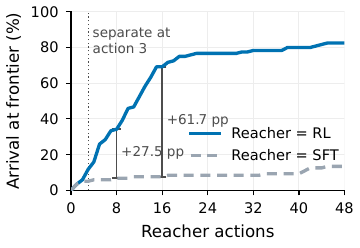}
  \vspace{-0.25cm}
  \caption{\textbf{Arrival at $\mathcal F_2$ by reacher action.} ALFWorld
  unseen split, $120$ paired trajectories per checkpoint.}
  \label{fig:overview}
  \vspace{-0.2cm}
\end{wrapfigure}
Figure~\ref{fig:overview} shows that the two checkpoints already diverge by
the third reacher action; by the end of the budget, RL reaches the solvable
frontier about six times as often as SFT. The curves record environment
membership in $\mathcal F_2$, the set of states exactly two valid actions from
success with enough budget to complete them (the general $\mathcal F_D$ is
defined in Section~\ref{sec:solvable-frontier}). This verified frontier gives
the common boundary for solver comparison in Section~\ref{sec:method}.
Endpoint success still combines arrival and completion, whereas reached-only
conditioning changes the target population; Section~\ref{sec:survivor} and
Appendix~\ref{app:robustness} quantify the resulting selection and budget
effects.

\subsection{Keeping Only Reached States Changes the Estimand}
\label{sec:survivor}

\begin{wrapfigure}{r}{2.4in}
  \vspace{-0.1cm}
  \centering
  \includegraphics[width=2.3in]{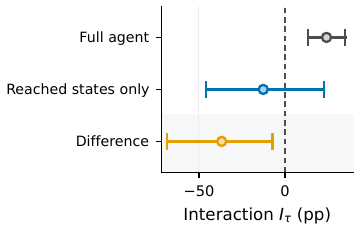}
  \vspace{-0.25cm}
  \caption{\textbf{$I_\tau$ under two estimands.} Same ALFWorld
  unseen-split episodes, scored two ways; $95\%$ paired bootstrap.}
  \label{fig:estimand}
  \vspace{-0.15cm}
\end{wrapfigure}
Figure~\ref{fig:estimand} evaluates the same ALFWorld episodes under two
different estimands. The full-population estimand keeps every initial episode.
An episode whose reacher never arrives has no continuation state and therefore
counts as a failure. This quantity combines how often a reacher arrives with
what a solver does after arrival. The reached-only estimand conditions on the
reacher having arrived and averages only over the states that this reacher
actually produced. It therefore measures conditional completion, not the
overall endpoint gain, and it does not compare a common set of states across
the two reachers.

The two estimands answer different questions. A reacher that arrives mainly on
easy tasks can have a different conditional solver gain from one that also
reaches harder tasks: the full-population estimand retains arrival frequency,
whereas reached-only normalizes it away. In our data the full-population
interaction is positive while the reached-only estimate has the opposite
direction and a wide interval; Appendix~\ref{app:robustness} quantifies the
support loss behind that interval.

Conditioning on arrival targets conditional completion, but using it to explain
the population-level RL gain induces selection because arrival depends on the
policy and task difficulty~\citep{hernan2004structural}. Restricting further to
episodes that both policies reach changes the target again. The protocol
therefore retains non-arrivals for the population endpoint and uses cloned
states for the conditional solver effect; the retention diagnostic appears in
Appendix~\ref{app:robustness}.

\section{Methodology: Checkpoint Handoff}
\label{sec:method}

\begin{figure}[t]
  \centering
  \includegraphics[width=0.94\linewidth]{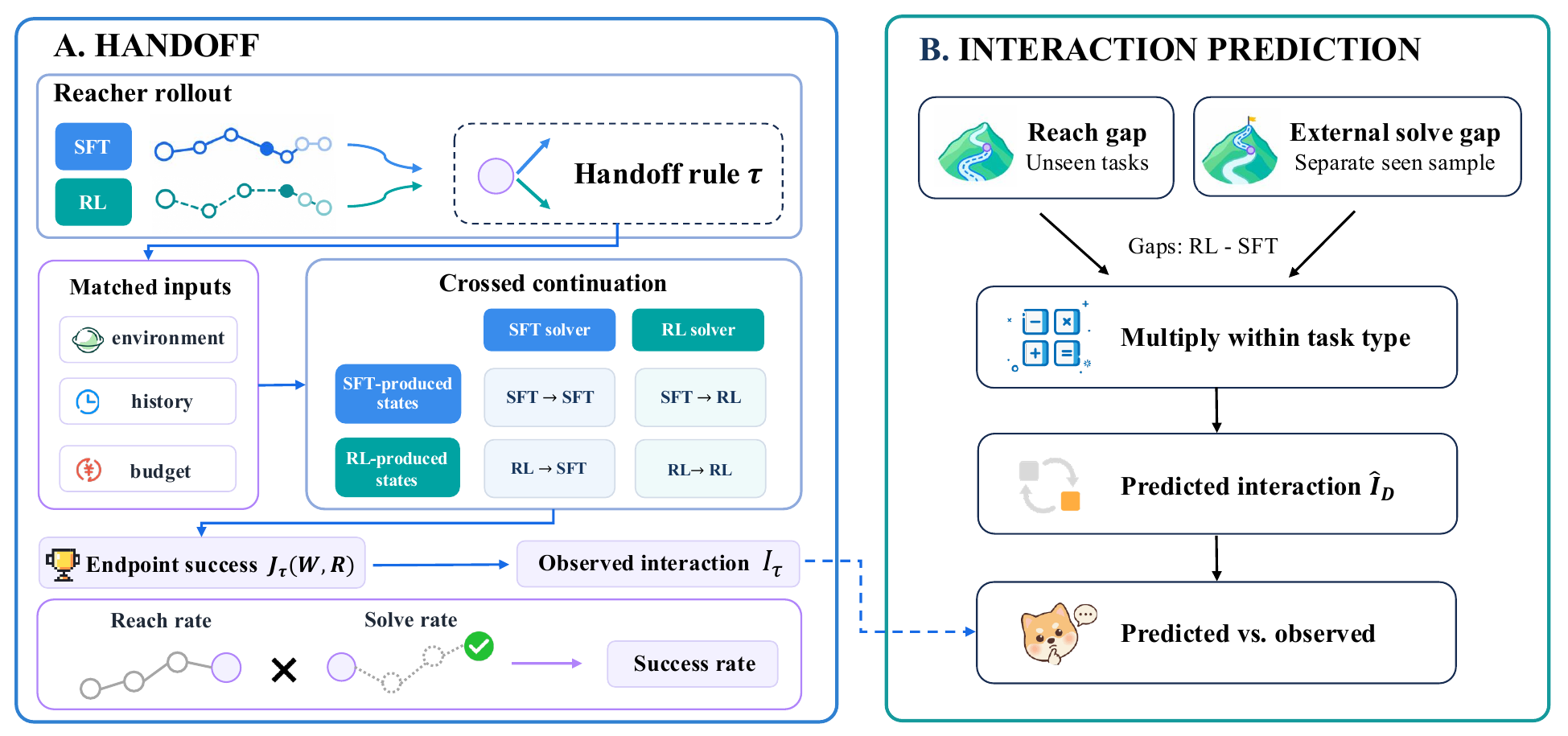}
  \caption{\textbf{Checkpoint handoff and interaction prediction.}
  \textbf{(A)} A benchmark-specific rule selects a handoff from each reacher's
  trajectory. Both solvers continue with matched inputs, yielding four endpoint
  success rates and their interaction. \textbf{(B)} On ALFWorld, target arrival
  gaps and independent \textsc{Solve} gaps give a predicted interaction,
  compared with the observed value (Equation~\ref{eq:composition-prediction}).}
  
  \label{fig:protocol}
\end{figure}

Checkpoint handoff crosses released reachers and solvers without retraining
(Equation~\ref{eq:reach-solve}). Solvers receive no label identifying the
reacher. Each solver continues live from the restored handoff state, so its
actions change the environment as they would in deployment.
Figure~\ref{fig:protocol} summarizes the design and the ALFWorld prediction
test; Appendix~\ref{app:reproducibility}, Table~\ref{tab:concerns-controls}
maps the controls to the quantities they identify.

\subsection{Handoff at Equal Remaining Distance}
\label{sec:solvable-frontier}

For replayable environments such as ALFWorld, we align handoffs by remaining
distance: after the same number of actions, one reacher may be one action
from success and another ten. Distance to success is used
as a goal-conditioned value~\citep{kaelbling1993learning}, a proximity measure
in embodied navigation evaluation~\citep{anderson2018evaluation}, and a
reward-shaping potential~\citep{ng1999policy}. We use it to index states, not
to reward them. For any valid action sequence
$\mathbf a$ from state $\sigma$, let $\Phi_T(\sigma,\mathbf a)$ be the state
obtained by exact replay and $|\mathbf a|$ its length. The remaining
distance, computed by replay in the environment, is
\begin{equation}
  \ell_T(\sigma)=\min_{\mathbf a\ \text{valid from}\ \sigma}\!\left\{
  |\mathbf a|:\ g_T\!\left(\Phi_T(\sigma,\mathbf a)\right)=1\right\},
  \label{eq:remaining-distance}
\end{equation}
with $\ell_T(\sigma)=\infty$ if no successful sequence exists. For $D\ge1$ the
solvable frontier is
\begin{equation}
  \mathcal F_D(T)=\left\{\sigma\in\mathcal S:\ \ell_T(\sigma)=D,\ b(\sigma)\ge D\right\},
  \label{eq:solvable-frontier}
\end{equation}
where $b(\sigma)$ is the budget component of $\sigma$. Membership depends only on the
environment, never on which checkpoint produced the state: every retained
state is exactly $D$ valid actions from success and has budget to get there.
With early budget $B_{\mathrm{early}}$, the handoff is the first hitting time
$\tau_D=\inf\{t\le B_{\mathrm{early}}:\Sigma_t\in\mathcal F_D(T)\}$, and an
empty set yields the non-arrival $\bot$ with $V_R(\bot)=0$. $\tau_D$ depends
on the reacher alone, as Section~\ref{sec:setting} requires; we write
$J_D=J_{\tau_D}$ and $I_D=I_{\tau_D}$. We use $D=B_{\mathrm{cont}}=2$, the
shortest horizon that lets a solver act, observe feedback, and act again.
Enumerating one- and two-action continuations certifies this distance against
the environment's success flag (Appendix~\ref{app:reproducibility}). Every
retained state requires two actions to succeed, and both solvers receive
exactly that budget.

\subsection{\textsc{Reach} and \textsc{Solve} from Cloned States}
\label{sec:cloned-state-cross}

Let $Z_W=\mathbf 1\{\Sigma_{\tau_D}\ne\bot\}$ indicate that reacher $W$
arrived. \textsc{Reach} is the arrival rate $A_W=\Pr(Z_W=1)$, the share of
episodes in which $W$ reaches the frontier at all. Non-arrivals are neither
discarded nor imputed. Conditional on arrival, and defined when $A_W>0$, let
$q_W^D$ be the natural frontier-state distribution, namely $d_W^{\tau_D}$
conditioned on $\Sigma_{\tau_D}\ne\bot$. We restore each reached state twice
and assign the SFT and RL solvers under identical environment,
history, admissible actions, budget, and seed schedule. We pair seeds because
seed choice alone moves reported agent performance by a wide
margin~\citep{henderson2018deep}. \textsc{Solve} is the
conversion rate $C_{W,R}^D=\mathbb E_{\Sigma\sim q_W^D}[V_R(\Sigma)]$, the
share of those arrivals that solver $R$ turns into success. With
$V_R(\bot)=0$ the endpoint is the product of the two rates,
\begin{equation}
  J_D(W,R)=A_W\,C_{W,R}^D,
  \label{eq:exact-factorization}
\end{equation}
so the reacher moves the first factor and the solver the second. Matching the
environment state and visible history fixes task progress and available
information; switching only the continuation checkpoint identifies the same-state
\textsc{Solve} gap
$\delta(\sigma)=V_{\mathrm{RL}}(\sigma)-V_{\mathrm{SFT}}(\sigma)$, and
\begin{equation}
  I_D=A_{\mathrm{RL}}\,\mathbb E_{\Sigma\sim q_{\mathrm{RL}}^D}[\delta(\Sigma)]
  -A_{\mathrm{SFT}}\,\mathbb E_{\Sigma\sim q_{\mathrm{SFT}}^D}[\delta(\Sigma)].
  \label{eq:interaction-decomposition}
\end{equation}
With $\bar\delta_W=\mathbb E_{\Sigma\sim q_W^D}[\delta(\Sigma)]$ the mean
\textsc{Solve} gap on states produced by reacher $W$,
\begin{equation}
  I_D=\underbrace{(A_{\mathrm{RL}}-A_{\mathrm{SFT}})\,\bar\delta_{\mathrm{RL}}}_{\text{arrival}}
  +\underbrace{A_{\mathrm{SFT}}\,(\bar\delta_{\mathrm{RL}}-\bar\delta_{\mathrm{SFT}})}_{\text{state}}.
  \label{eq:interaction-channels}
\end{equation}
The arrival term is positive whenever RL both reaches and solves better. The
state term is positive only if the RL solver's advantage grows on RL-produced
states. For a prespecified set of states
$\mu_D$, chosen independently of solver assignment and outcomes, we also
report
$\Delta_{\mathrm{solve}}(\mu_D)=\mathbb E_{\Sigma\sim\mu_D}[\delta(\Sigma)]$.
We report $\delta$ separately on SFT-produced and RL-produced states, so a
\textsc{Solve} gain cannot be an artefact of RL states being easier.

\subsection{Predicting Interaction from Reach and Solve Gaps}
\label{sec:external-composition}

Equation~\ref{eq:exact-factorization} is an identity on the same natural
states. A stronger test forecasts the interaction from components measured
elsewhere, before the target continuations are run. Let $k\in\mathcal K$ index
task strata with prevalence $p_k$. Natural rollouts on the target split give
the arrival rate $A_W(k)=\Pr(Z_W=1\mid k)$. An independently selected
set of $\mathcal F_D$ states gives each solver's \textsc{Solve} rate
$C_R^{\mathrm{ext}}(k)$ on states the target checkpoints never visited; its
selection and scoring never touch target continuation outcomes.
Under this component model, the interaction is the prevalence-weighted
product of the two gaps,
\begin{equation}
  \widehat I_D^{\mathrm{comp}}=\sum_{k}p_k\,
  \big[A_{\mathrm{RL}}(k)-A_{\mathrm{SFT}}(k)\big]
  \big[C_{\mathrm{RL}}^{\mathrm{ext}}(k)-C_{\mathrm{SFT}}^{\mathrm{ext}}(k)\big].
  \label{eq:composition-prediction}
\end{equation}
The residual $R_{\mathrm{comp}}=I_D-\widehat I_D^{\mathrm{comp}}$ is a
forecast error: nothing in
Equation~\ref{eq:composition-prediction} is estimated from the observed
$I_D$. Its transport condition, in the sense of
\citet{pearl2011transportability}, is that within $k$ the external rate
approximates continuation success on naturally reached target states
regardless of their reacher. The per-type residuals quantify how well this
condition holds across the target task distribution.

\section{Experiments}
\label{sec:experiments}

\subsection{Setup}
\label{sec:experimental-design}

TravelPlanner~\citep{xie2024travelplanner} tests RQ1 across model scales
using recorded-history handoffs. ALFWorld~\citep{shridhar2021alfworld}
supports verified-frontier handoffs for RQ2 and RQ3. Agent-STAR~\citep{wu2026agentstar}
and SkillRL~\citep{xia2026skillrl} are independent Qwen-family training
pipelines. Decoding settings were identical within each SFT/RL pair and
fixed before any handoff ran (Appendix~\ref{app:reproducibility}).

\paragraph{TravelPlanner and Agent-STAR.}
We cross released 1.5B, 3B, and 7B Agent-STAR SFT/RL checkpoints on the same
$180$ TravelPlanner validation tasks; their immutable revisions are listed in
Appendix~\ref{app:reproducibility}, Table~\ref{tab:model-revisions}.
TravelPlanner's two-stage setting separates information collection from plan
generation over a static database~\citep{xie2024travelplanner}. Agent-STAR
records tool interactions in a message history and terminates with a submitted
plan~\citep{wu2026agentstar}. We therefore use the recorded trajectory as the
handoff input: task prompt, model messages, tool calls, and tool responses.
Each reacher runs to termination under the native $60$-decision limit; we cut
immediately before its final model call. This retains the reacher's accumulated
information and intermediate reasoning while leaving plan completion to the
solver. The cut depends only on the recorded reacher run. We reset the task,
replay the prefix, and verify identical reconstructed histories. Both solvers
receive the same record and at most $B=4$ model decisions, allowing further
tool use before submission under a common budget. A fixed Formatter converts
the result for the official evaluator. Crossing reachers and solvers gives four
assignments, each evaluated on the same 180 tasks, or 720 rows per scale
(Appendix~\ref{app:reproducibility}).

\paragraph{ALFWorld and SkillRL.}
ALFWorld's TextWorld engine represents household states and action effects
symbolically~\citep{shridhar2021alfworld}. We use its admissible actions and
success flag to verify remaining distance by enumeration. The unseen split places tasks in rooms
held out from training, and the seen split reuses training rooms. We run
released SkillRL 7B SFT/RL checkpoints on $30$ prespecified unseen-split tasks, five from each official
task type, with four paired seeds. \textsc{Reach} comes from natural
rollouts: a reacher acts for at most $48$ steps, and after each action the
environment is asked whether the current state lies on $\mathcal F_2$
(Section~\ref{sec:solvable-frontier}).
\textsc{Solve} comes from replay. We reset the task under the same seed and
replay the recorded reacher actions to restore the arrival state. Observation,
admissible actions, and rendered prompt must match the recording. Each solver
then takes two actions from that clone, in a separate replay of the same
state. This yields $240$ natural trajectories, $115$
verified frontier states, and $480$ endpoint rows, scored by the
environment's native \texttt{won} flag. A separately selected seen-split sample of the same
design supplies the external \textsc{Solve} profile
(Appendix~\ref{app:external-panel}) and a replication. All replay,
solvability, pairing, prompt, and completeness checks pass
(Appendix~\ref{app:reproducibility}).

\paragraph{Inference.}
Arrival and endpoint rates are empirical means over prespecified tasks and
paired seeds, with zero endpoint success for every non-arrival. All
uncertainty uses $100{,}000$ task-level paired bootstrap
resamples~\citep{efron1994introduction} that keep seeds, the four checkpoint
assignments, and task-type counts together. Tasks are the unit that varies,
and few-run evaluations call for interval estimates rather than point
scores~\citep{agarwal2021deep}. For the component-based prediction, each draw
independently samples tasks within each task type from the target and external
samples, retains all paired seeds for every sampled task, and recomputes the
observed interaction, predicted interaction, and residual from that draw.
On ALFWorld, same-state contrasts identify the solver effect because the
environment decides $\tau_D$ before any solver runs, both solvers start from
exact clones, and only the solver changes.

\subsection{RQ1: The Solver Gain from SFT to RL Depends on the Reacher}
\label{sec:travelplanner-interaction}

\begin{figure}[!htbp]
  \centering
  \includegraphics[width=\linewidth]{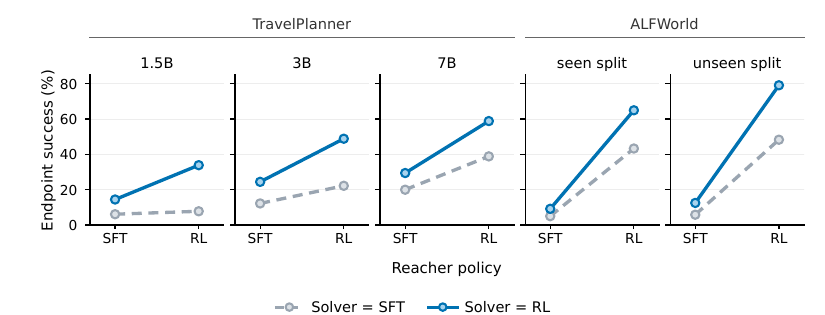}
  \caption{\textbf{Endpoint success for the four reacher by solver assignments.}
  Non-parallel lines indicate a reacher--solver interaction.}
  
  \label{fig:endpoint-factorial}
\end{figure}

\begin{table}[!htbp]
  \centering
  \footnotesize
  \caption{\textbf{Endpoint success for the four reacher ($W$) by solver ($R$)
  assignments.} Subscripts on the $R$=RL columns give the solver effect under
  that reacher, and the last column is their difference,
  $I_\tau$ of Equation~\ref{eq:factorial-interaction}. Brackets are $95\%$
  paired-bootstrap intervals.}
  
  \label{tab:endpoint-factorial}
  \input{tables/table1_endpoint_factorial}
\end{table}

\paragraph{The Solver Effect Is Two to Five Times Larger Under the RL Reacher}
Table~\ref{tab:endpoint-factorial} compares the SFT and RL solvers separately
under each reacher. In Figure~\ref{fig:endpoint-factorial} the vertical gap
between the two lines is the solver effect, so a gap that changes across
reachers indicates an interaction. The interaction is positive in all five
conditions, and every
lower bound in Table~\ref{tab:endpoint-factorial} lies above zero. The size
of this gap is the informative part. On TravelPlanner, switching to
the RL solver gains between two and three times as much on RL-produced
histories as on SFT-produced ones, at every scale. On ALFWorld the ratio
rises to about five. The SFT reacher rarely hands the solver a finishable
state, and no SFT-reacher entry exceeds $12.5\%$ whichever solver continues. A
better solver pays off only on episodes that reach a solvable state, and on
ALFWorld the larger gain comes entirely from arrival
(Sections~\ref{sec:alfworld-reach}--\ref{sec:alfworld-composition}).

\paragraph{With Scale, the Gain Moves From the Solver to the Reacher}
At 1.5B the SFT reacher with the RL solver beats the RL reacher with the SFT
solver, $14.4\%$ against $7.8\%$; an RL history is worth little to an SFT
solver. At 7B the order flips, $29.4\%$ against $38.9\%$, so the RL history
alone now carries most of the gain. The interaction shrinks from 1.5B to 7B,
but the three intervals overlap and the 7B lower bound sits close to zero.
We read the decline as an observation, not a trend. The intervals support a
positive interaction at every scale and under two independently trained
pipelines.

\subsection{RQ2a: RL Reaches the Frontier More Often than SFT}
\label{sec:alfworld-reach}

\paragraph{RL Reaches an Equally Solvable State Six Times as Often}
On ALFWorld each entry of Table~\ref{tab:endpoint-factorial} equals the
arrival rate in Table~\ref{tab:reach-solve}(a) times the completion rate in
Table~\ref{tab:reach-solve}(b), to the reported precision.
\begin{wrapfigure}{r}{2.4in}
  \vspace{-0.35cm}
  \centering
  \includegraphics[width=2.3in]{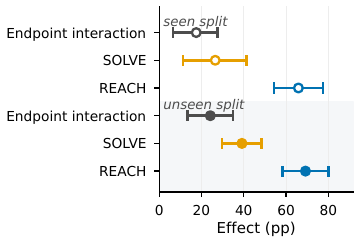}
  \vspace{-0.25cm}
  \caption{\textbf{\textsc{Reach}, \textsc{Solve}, and the endpoint interaction.}
  Filled markers are the unseen split, open the seen split.}
  \label{fig:components}
  \vspace{-0.2cm}
\end{wrapfigure}
Equation~\ref{eq:reach-solve} is an identity here, and the question is how
much of the gain sits in each term. On the unseen split the SFT reacher
reaches $\mathcal F_2$ in $13.3\%$ of trajectories and the RL reacher in
$82.5\%$. Every frontier state is exactly two actions from success, so the
contrast counts arrivals at a common boundary. Mean actions to
arrival or stopping, $44.0$ against $17.8$, separate two regimes: the SFT
reacher usually runs to the $48$-action limit without arriving, the RL
reacher usually arrives within the first third of the budget. The gap
replicates on the seen split, $15.8\%$ against $81.7\%$, and \textsc{Reach}
is the component on which the two splits agree most closely
(Figure~\ref{fig:components}). Appendix~\ref{app:robustness} reports the
split replication, budget contrast (Figure~\ref{fig:budget} and
Table~\ref{tab:budget}), and non-arrival audit: RL with eight
actions already out-reaches SFT with forty-eight, and almost every SFT
non-arrival is a loop that repeats actions without changing the state. More
budget does not close the deficit.

\begin{table}[!htbp]
  \centering
  \footnotesize
  \caption{\textbf{\textsc{Reach} and \textsc{Solve} on ALFWorld.} Each row is a
  reacher. (a) Arrival at $\mathcal F_2$ over $120$ trajectories, with mean
  actions to arrival or stopping. (b) Completion by each solver from identical
  replayed states that the reacher produced. Brackets are task-stratified
  paired-bootstrap $95\%$ intervals; a dash marks a stratum too small for one.}
  \label{tab:reach-solve}
  \input{tables/table2_reach_solve}
\end{table}

\subsection{RQ2b: RL Outperforms SFT from Identical States}
\label{sec:alfworld-solve}

\begin{table}[!htbp]
  \centering
  \footnotesize
  \caption{\textbf{\textsc{Solve} split into choosing a bridge action and finishing
  from it.} Shares of reached states with $95\%$ bootstrap intervals; the last
  two columns count states completed by only one solver ($n=115$
  on the unseen split, $117$ on the seen split).}
  \label{tab:action-stage}
  \input{tables/table5_action_stage}
\end{table}

\paragraph{The RL Solver Is Near Ceiling Whoever Produced the State}
From two exact clones of each of the $115$ reached unseen-split
states, the SFT solver completes $56.5\%$ and the RL solver $95.7\%$
(Table~\ref{tab:reach-solve}b). The clones share task, history, admissible
actions, two-action budget, and seed schedule, so the gap is a solver effect,
and its lower bound lies well above zero on both splits. The origin of the
state matters only to the SFT solver: it completes $43.8\%$ of the states its
own reacher produced and $58.6\%$ of the RL-produced ones, while the RL solver
completes $93.8\%$ and $96.0\%$. The same-state effect is no smaller on
SFT-produced states ($50.0$ against $37.4$ points), so the \textsc{Solve} gain
is not an artefact of RL states being easier. Of the $115$ states, $45$ are won by the RL solver only and none by
the SFT solver only (Table~\ref{tab:action-stage}). The same table splits
\textsc{Solve} into choosing a bridge action and finishing from it. SFT loses
at both stages, and the RL advantage at the second action is the less stable
across splits. Appendix~\ref{app:robustness} gives the corresponding
action-stage breakdown.
Appendix~\ref{app:case-study} (Table~\ref{tab:case-study}) walks through one
state: holding the pencil,
the SFT solver spends its first action looking around while the RL solver
walks to a shelf and places it.

\subsection{RQ3: Predicting Reacher--Solver Interaction from Reach and Solve Gaps}
\label{sec:alfworld-composition}

\begin{wrapfigure}{r}{2.4in}
  \vspace{-0.35cm}
  \centering
  \includegraphics[width=2.3in]{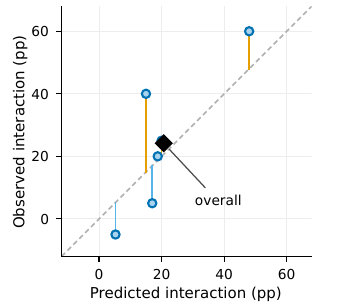}
  \vspace{-0.25cm}
  \caption{\textbf{Predicted against observed interaction, per task type.}
  The dashed line is $y=x$; the black diamond is the unweighted mean across
  the six task types.}
  \label{fig:recomposition}
  \vspace{-0.2cm}
\end{wrapfigure}
\paragraph{Independent Reach and Solve Gaps Recover the Aggregate Interaction}
We test whether the interaction can be recovered from its two component gaps.
For each task type, we multiply the RL--SFT difference in frontier arrival by
the RL--SFT difference in continuation success, then average across the six
balanced types. Arrival rates come from the target split, while continuation
success is measured on an independent seen-split sample with no state or
continuation outcome shared with the $115$ replayed target states. The target
\texttt{valid\_unseen} states are reserved for evaluating the observed
interaction; the independent \texttt{valid\_seen} sample supplies the external
\textsc{Solve} profile, making this an out-of-sample transport test across
splits conditional on task type.
Equation~\ref{eq:composition-prediction} predicts an interaction of $20.7$
percentage points, compared with an observed value of $24.2$ points
(Figure~\ref{fig:recomposition}, Appendix~\ref{app:external-panel}). The
black-diamond overall point is the unweighted mean of the six task-type
predictions and observations, $(20.7,24.2)$. The prediction recovers the
positive interaction and its aggregate scale, with a residual of $3.5$ points
(95\% CI $[-7.2,14.2]$). Equation~\ref{eq:interaction-channels} agrees in
sample, with arrival terms of $25.9$ and $17.5$ points and state terms of
$-1.7$ and $0.0$ on the unseen and seen splits, so the interaction on ALFWorld
is carried by arrival. Appendix~\ref{app:external-panel} gives the
per-type comparison, including Figures~\ref{fig:task-type-solve} and
\ref{fig:task-type-recomposition}, along with their source rates in
Table~\ref{tab:recomposition}.

\section{Conclusion}
\label{sec:conclusion}

An agent's earlier actions shape the environment states it later acts from, so
its final success rate mixes two abilities: reaching useful states and
completing the task once there. In this paper, we propose checkpoint handoff,
an evaluation protocol that separates the two by letting one checkpoint reach
a state and another continue from an exact copy of it. Experiments with SFT
and RL checkpoints show that the RL solver's advantage is larger when RL is
also the reacher. On ALFWorld, RL improves both abilities, and this larger
advantage comes from reaching solvable states more often.
Looking beneath final success in this way gives a clearer picture of what
agentic RL changes.

\section*{AI Use Statement}

This study uses AI tools to write parts of the code, refine the idea, and adjust the LaTeX formatting.

\section*{Reproducibility Statement}

Appendix~\ref{app:reproducibility} lists the checkpoints and revisions, task
samples, generation settings, and replay checks.
Appendix~\ref{app:external-panel} describes the independent \textsc{Solve}
sample, and Appendix~\ref{app:robustness} gives the seen-split replication and
additional analyses. Code is available at
\url{https://github.com/Xuanxuana1/AgenticRL_Handoff}. The
supplementary material includes the configurations, continuation records,
Formatter replies, rewards, and summary statistics used in the paper.

\bibliography{references}
\bibliographystyle{iclr2027_conference}
\clearpage

\clearpage
\appendix

\section{Related Work}
\label{app:related}

\paragraph{Training and comparing SFT and RL agents.}
Multi-turn RL for agents optimizes whole trajectories. RAGEN treats the
rollout as the unit of optimization and studies when training collapses, Agent
Lightning decouples agent execution from training and assigns credit at the
transition level, and AgentGym-RL lengthens the interaction horizon during
training~\citep{wang2025ragen,luo2025agentlightning,xi2025agentgymrl}.
SkillRL grows a recursive skill library, and Agent-STAR distills a training
recipe for long-horizon tool use~\citep{xia2026skillrl,wu2026agentstar}.
Diagnostics of such runs localize reasoning collapse and token-level action
bottlenecks~\citep{wang2026ragen2,he2026actfocus}. Studies that compare the
two regimes report that RL generalizes where SFT memorizes, that RL
fine-tuning restores out-of-distribution ability lost during SFT, and that SFT
generalization depends on data and model
capability~\citep{chu2025sft,jin2025heals,ren2026rethinking}. These works
measure whether and how much RL helps, each scoring a checkpoint on the states
it reaches by itself. We take released checkpoints from two of these pipelines
and ask which ability the gain comes from.

\paragraph{Agent evaluation beyond endpoint success.}
AgentBoard scores progress toward subgoals, process evaluation grades each
step of an agentic trajectory, and proxy-state rewards verify the environment
state a tool call leaves
behind~\citep{ma2024agentboard,gritta2026process,chuang2026proxy}.
Exploration analyses measure how often a policy enters useful regions, guide
it there with action-level supervision, or assign credit across several
timescales of environment
feedback~\citep{ye2026look,ji2026actguide,huo2026environmental}. Failure
diagnosis separates errors made before a capability becomes exercisable from
errors made once it is~\citep{shao2026beyond}. Each of these describes the
rollout of a single policy, so they either restrict attention to the states
that policy reached (Section~\ref{sec:survivor}) or change the state
distribution and the continuation value together
(Section~\ref{sec:two-factors}). Checkpoint handoff adds the comparison they
lack: a second policy acting from the same state.

\paragraph{Counterfactual attribution of agent outcomes.}
Counterfactual effect decomposition attributes multi-agent outcomes to
individual actions and environment
transitions~\citep{triantafyllou2025counterfactual}, and perception-aware
rewards separate vision-language errors of perception from errors of
reasoning~\citep{wang2026badseeing}. Both decompose the outcome of a single
model into parts of the process that produced it, whereas checkpoint handoff
compares two policies acting from the same state. We are also aware of
concurrent work on per-step model routing, which forks live agent trajectories
and shows that predicting the outcome of a mid-episode model switch from each
model's logged runs misjudges it~\citep{gonuguntla2026replaygap}. That work
evaluates routers for efficiency, while ours focuses on what agentic RL
training changes in a policy. Our claims concern behavior, not internal
mechanisms.

\section{Reproducibility Details}
\label{app:reproducibility}

\paragraph{Released checkpoints.}
We evaluate existing checkpoints without retraining them.
Table~\ref{tab:model-revisions} lists the immutable revisions used throughout
the paper. Agent-STAR~\citep{wu2026agentstar} provides three SFT/RL pairs for
TravelPlanner, and SkillRL~\citep{xia2026skillrl} provides one SFT/RL pair for
ALFWorld. The two releases constitute
independent training pipelines, but both use architectures from the broad
Qwen family; the experiment therefore does not establish transfer across
unrelated base-model architectures.

\begin{table}[H]
  \centering
  \scriptsize
  \caption{Released checkpoints and fixed revisions.}
  \label{tab:model-revisions}
  \begin{tabular}{@{}p{0.14\textwidth}p{0.31\textwidth}p{0.46\textwidth}@{}}
    \toprule
    Benchmark & Checkpoint & Revision \\
    \midrule
TravelPlanner & \nolinkurl{xxwu/Agent-STAR-SFT-1.5B} & \nolinkurl{b3d55c10e763231d13f591bfa2ad2cb99d19b981} \\
    TravelPlanner & \nolinkurl{xxwu/Agent-STAR-RL-1.5B}  & \nolinkurl{279b7c75586b2a6ada3eac5a5f406a147522c787} \\
    TravelPlanner & \nolinkurl{xxwu/Agent-STAR-SFT-3B}   & \nolinkurl{dfb0d943ad8c7a1d6a5e38064a75a6f4e8b4e503} \\
    TravelPlanner & \nolinkurl{xxwu/Agent-STAR-RL-3B}    & \nolinkurl{705723c62e82ed15844783f221572ea1c78f4185} \\
    TravelPlanner & \nolinkurl{xxwu/Agent-STAR-SFT-7B}   & \nolinkurl{945532c2ca454caaf1f62d1c42916cbec0a2973a} \\
    TravelPlanner & \nolinkurl{xxwu/Agent-STAR-RL-7B}    & \nolinkurl{7013d9fd0fff692d8f6f352ae5db4351d174eab9} \\
    ALFWorld & \nolinkurl{Jianwen/Alfworld-7B-SFT} & \nolinkurl{ba9c962eef80a49fc63a94c9728209a36057c671} \\
    ALFWorld & \nolinkurl{Jianwen/Alfworld-7B-RL}  & \nolinkurl{2ce16cb90e6357892dde201928279d4513d35c59} \\
    \bottomrule
  \end{tabular}
\end{table}

\paragraph{TravelPlanner protocol.}
For each scale, Val180 contains 60 easy, 60 medium, and 60 hard tasks. The
primary analysis crosses two reachers and two solvers at budget $B=4$,
yielding 720 rows per scale. Inference uses Python 3.10.20, PyTorch 2.6.0
with CUDA 11.8, vLLM 0.8.1 with CUDA 11.8, FP16, and a 32K-token context.
We use nucleus sampling~\citep{holtzman2020curious} with temperature 0.6,
top-$p$ 0.95, and seed 2027. Confidence intervals use
100,000 task-level paired bootstrap resamples. The archived Formatter alias is
\texttt{deepseek-chat}; its raw replies are retained. All requests completed,
and recomputing the official Agent-STAR evaluator produced zero per-example
reward differences. The three formal terminal summaries pass their
protocol acceptance checks.

\paragraph{TravelPlanner handoff construction.}
Each reacher first runs its released checkpoint under the native 60-decision
and 60-tool limits. The handoff is the point immediately before that
trajectory's final model call, rather than a common elapsed step. The rule is
a function of the recorded reacher trajectory alone, so it satisfies
Section~\ref{sec:setting}, but its states are late rather than early. We reset the
task and replay every preceding assistant action and tool response, then
require the reconstructed message list to equal the recorded list. For each
history, the SFT and RL solvers receive that entire list and the recorded
final-turn seed; no message is summarized, reordered, or semantically
repaired. The recorded first solver decision is the $B=1$ snapshot. If it is a
valid tool call, the evaluator executes it, appends its native tool response,
and permits at most three further decisions, using consecutive paired seeds,
to form $B=4$. Answers, invalid actions, and context limits are absorbing. Thus
$B=4$ counts four post-handoff model decisions, not the four
reacher--solver assignments. The
Formatter is applied only after the solver terminates. At every scale, the
$B=4$ audit contains 720 unique rows, 180 per assignment, with equal paired
input
hashes, no Formatter request failures, and no official-score recomputation
differences.

\paragraph{Formatter measurement audit.}
The fixed-handoff Val180 run initially contained two unique empty Formatter
envelopes, corresponding to four endpoint rows. The separately registered
repair protocol retried each input exactly once with the same model, prompt,
JSON contract, and disabled-thinking setting, using no new trajectories and
changing no checkpoint assignment; it writes to an isolated artifact. Both
requests returned valid envelopes. The repaired directory contains 292 valid
unique Formatter samples, 720 formatted rows, and 720 officially scored rows;
the evaluator recomputation reported zero mismatches. On the 716 unaffected
rows, all identity fields, formatted-plan hashes, official metrics, and
$r_{\mathrm{sum}}$ are identical to the source artifact. Eight diagnostic
\texttt{reward\_details} messages differ only in the order of an unordered
list, which is recorded separately and does not change any score. The
fixed-handoff interaction remains $7.22$ points with 95\% CI $[1.67,12.78]$,
so the positive interaction is not explained by the two transient measurement
failures.

\paragraph{ALFWorld protocol.}
The unseen-split sample (\texttt{valid\_unseen} in the ALFWorld release; the
seen-split sample is \texttt{valid\_seen}) contains 30 tasks stratified across the six
ALFWorld task types. Each released checkpoint receives 48 early environment steps. Natural
arrival is measured at the environment-verified distance-two frontier
$\mathcal F_2$. Every arriving state is replayed, verified, cloned, and
continued by both solvers with two remaining environment steps and four paired
seed slots. Inference uses vLLM 0.8.1, BF16, a 4,096-token prompt limit, a
512-token response limit. We use temperature-scaled multinomial sampling with
temperature 0.4 and top-$p$ 1 (no probability truncation), with selection seed
2029.
The outcome is the environment's native \texttt{won} flag. The main protocol
produces 115 verified frontier states and passes all replay, pairing, and
completion checks.

\paragraph{ALFWorld solvability check and clone audit.}
For a candidate action prefix, the check first rejects terminal states,
then replays every currently admissible action. It rejects the candidate if any
first action wins; otherwise it replays every admissible second action from
each nonterminal successor and accepts only if at least one two-action path
wins. This exhaustive local search includes alternative shortest paths and
states from which some actions are irreversible. Across the 115 accepted
handoffs, it tested 3,227 first actions and 89,511 second actions, and never
failed.

The environment is restored by deterministic reset with the same task seed and
exact replay of the recorded reacher actions; we do not serialize a Python
environment object. After reset and every reacher action, equality is checked
on a JSON-safe record containing the observation, ordered admissible-action
list, score, terminal flag, and \texttt{won} flag. At handoff, the visible
history, rendered prompt, prompt-token count, and token-ID digest must also
agree. The two solvers are run in separate replays from this operational
clone. All 230 arrival-solver replays passed, all 115 solver pairs had equal
inputs and seed schedules, and no prompt exceeded 4,096 tokens.
Table~\ref{tab:concerns-controls} maps each common attribution concern to the
protocol control that defines the reported quantity.

\begin{table}[t]
  \centering
  \scriptsize
  \setlength{\tabcolsep}{4pt}
  \renewcommand{\arraystretch}{1.08}
  \caption{\textbf{Design questions and the protocol control that answers each.}}
  
  \label{tab:concerns-controls}
  \begin{tabular}{@{}p{0.20\textwidth}p{0.42\textwidth}p{0.30\textwidth}@{}}
    \toprule
    Design question & Protocol control & Reported quantity \\
    \midrule
    Are states equally close to success?
      & Exhaustive $\mathcal F_2$ replay check that never consults a
        checkpoint, and a matched two-action solver budget
      & Arrival at a common solvable frontier \\
    Do solvers receive the same input?
      & Exact replay with equal environment observations, history, admissible
        actions, budget, prompt, and paired seeds
      & Same-state \textsc{Solve} effect \\
    Are difficult non-arrivals removed?
      & Every natural non-arrival remains a zero endpoint outcome; reached-state
        results are reported separately
      & Deployed-agent endpoint interaction \\
    Does scoring introduce the interaction?
      & Environment-native \texttt{won} in ALFWorld; archived Formatter replies
        and exact official-score recomputation in TravelPlanner
      & Stable benchmark outcomes \\
    Is the interaction prediction fitted to the target split?
      & Conversion is measured on an independently selected sample before it
        is combined with target arrival
      & Out-of-sample interaction prediction \\
    \bottomrule
  \end{tabular}
\end{table}

\section{Independent Solve Profile Sample}
\label{app:external-panel}

The external \textsc{Solve} profile is selected before inspecting continuation
outcomes on the target \texttt{valid\_unseen} sample. It comes from a separate
\texttt{valid\_seen} experiment containing five tasks per task type. Four
paired seeds yield 120 distance-two states per solver. We combine these six
type-specific conversion rates with target-split type-specific arrival rates.
This estimator requires transportability of continuation success within task
type from the independent sample to the target split. It deliberately does not
assume that task type is sufficient for every aspect of state difficulty.

\begin{figure}[t]
  \centering
  \includegraphics[width=0.78\linewidth]{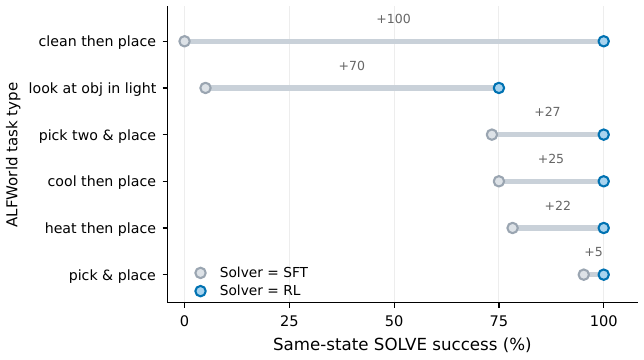}
  \caption{\textbf{Same-state \textsc{Solve} by task type on \texttt{valid\_unseen}.}
  Rows are sorted by the RL-minus-SFT gap. The source data report no
  per-type intervals.}
  
  \label{fig:task-type-solve}
\end{figure}

\begin{figure}[t]
  \centering
  \includegraphics[width=\linewidth]{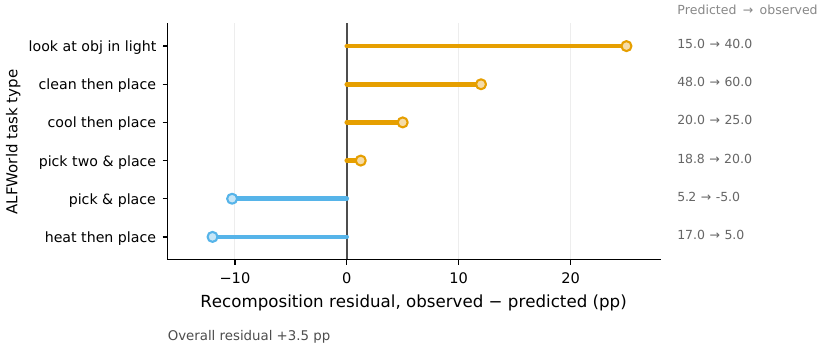}
  \caption{\textbf{Per-task-type interaction-prediction residual, observed minus predicted.}
  Positive means the external profile underpredicts. The source data report
  no per-type intervals.}
  
  \label{fig:task-type-recomposition}
\end{figure}

The resulting interaction prediction is 20.7 percentage points, compared with
an observed interaction of 24.2 points. The residual is 3.5 points with 95\%
CI $[-7.2,14.2]$, quantifying the remaining difference between prediction and
observation and the uncertainty from the finite task samples.
Figure~\ref{fig:task-type-solve} plots both solver levels by task type rather
than their difference: \emph{pick \& place} gains only 5 points because the
SFT solver already completes 95.2\% of those states, whereas \emph{clean then
place} gains 100 points from a floor of zero.
Figure~\ref{fig:task-type-recomposition} shows the per-type residuals behind
the 3.5-point aggregate: type-level errors are larger and have both signs.
Table~\ref{tab:recomposition} lists the same residuals
with the arrival and conversion rates they are computed from.
Type-level residuals run from $+25.0$ on look at object in light to $-12.0$
on heat then place and cancel in the unweighted mean. Comparing the external
conversion gap with the same-state \textsc{Solve} gap on the target's own
replayed states (Figure~\ref{fig:task-type-solve}) locates the error. For
look at object in light the external profile gives a $25$-point gap where
the target states show $70$, and the prediction falls $25$ points short. For
pick and place it gives $15$ where the target shows $4.8$, and it overshoots
by $10.2$. Five of the six residuals share the sign of this difference; heat
then place is the exception, with gaps of $20$ and $21.7$ that agree. The
error sits in the transport of the within-type \textsc{Solve} profile, which
is the assumption the external test was designed to expose.

\begin{table}[t]
  \centering
  \footnotesize
  \caption{\textbf{Per-task-type interaction prediction
  (Equation~\ref{eq:composition-prediction}).} Arrival is from the target
  \texttt{valid\_unseen} split and the \textsc{Solve} profile from the
  independent sample. The overall row is the unweighted mean over types, the
  stratified bootstrap unit; dashes mark cells it leaves undefined.}
  
  \label{tab:recomposition}
  \input{tables/table4_recomposition}
\end{table}

\section{Robustness Checks}
\label{app:robustness}

\paragraph{Seen-split replication.}
On a separately selected, type-balanced 30-task \texttt{valid\_seen} sample, RL
improves frontier arrival by 65.8 points with 95\% CI $[54.2,77.5]$, improves
same-state completion by 26.5 points with CI $[11.2,41.3]$, and yields a
positive endpoint interaction of 17.5 points with CI $[6.7,27.5]$. All three
effects replicate on a different split with positive lower confidence bounds.

\paragraph{\textsc{Solve} by solver action.}
Every frontier state admits bridge actions, first actions from which one
further action wins, so \textsc{Solve} factors into choosing a bridge action
and finishing from it (Table~\ref{tab:action-stage}). On
\texttt{valid\_unseen} the SFT solver chooses a bridge action on $72.2\%$ of
states and finishes from one on $78.3\%$ of those; the RL solver does so on
$97.4\%$ and $98.2\%$. The two stages contribute comparably to the SFT
deficit, $25.2$ and $19.9$ points, and their products reproduce the
completion rates of $56.5\%$ and $95.7\%$. Of the $115$ states, $45$ are
won by the RL solver only and none by the SFT solver only. On
\texttt{valid\_seen} the RL solver alone wins $41$ states and the SFT solver
alone wins $10$. The
SFT solver behaves almost identically on the two splits, $72.6\%$ against
$72.2\%$ at the bridge stage and $49.6\%$ against $56.5\%$ overall. The
change is on the RL side: its finishing rate given a bridge action falls
from $98.2\%$ to $82.4\%$, while its bridge rate falls only from $97.4\%$ to
$92.3\%$. The replication preserves the direction and the two-stage
structure of the \textsc{Solve} effect; the RL advantage at the second
action is the less stable of the two.

\paragraph{Early-action budget diagnostic.}
We reuse archived \texttt{valid\_seen} trajectories at common early-action
budgets 8, 16, and 48. Table~\ref{tab:budget} lists the arrival levels and
their contrasts. This post-hoc diagnostic shows a positive RL arrival advantage
at eight steps that remains large as both checkpoints receive more steps. In
the direct cross-budget comparison, RL at eight actions exceeds SFT at 48
actions by 31.7 points with CI $[19.2,44.2]$. Budget data exist for
\texttt{valid\_seen} only; no accepted \texttt{valid\_unseen} budget artefact
exists, so these intervals are not attached to the \texttt{valid\_unseen}
curve of Figure~\ref{fig:overview}. The claim is bounded by the protocol:
cumulative arrival within 48 reacher actions under frozen policies, with no
extrapolation to additional sampling, search, or replanning.

\begin{figure}[t]
  \centering
  \includegraphics[width=3.0in]{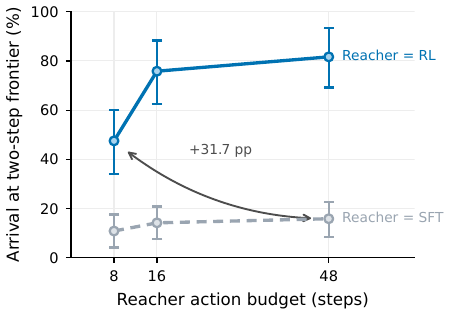}
  \caption{\textbf{Frontier arrival against reacher action budget on
  \texttt{valid\_seen}.} Arrival at three budgets on the separately selected
  \texttt{valid\_seen} sample, with task-stratified paired-bootstrap $95\%$
  intervals. The \texttt{valid\_unseen} curve is Figure~\ref{fig:overview}.}
  
  \label{fig:budget}
\end{figure}

\begin{table}[t]
  \centering
  \footnotesize
  \caption{\textbf{Frontier arrival by reacher budget on \texttt{valid\_seen}.} The
  bolded row contrasts RL at 8 actions with SFT at 48. Brackets are
  task-stratified paired-bootstrap 95\% intervals from $100{,}000$ resamples.}
  
  \label{tab:budget}
  \input{tables/table3_budget}
\end{table}

\paragraph{Non-arrival failure modes.}
Every \texttt{valid\_unseen} trajectory that never reaches $\mathcal F_2$
($104$ SFT, $21$ RL, all running the full $48$ actions) is assigned one of
seven mutually exclusive labels by an automatic audit of its action sequence.
For SFT, $102$ trajectories are loops that repeat actions without changing
the state and $2$ miss a subgoal; for RL, $18$ are loops, $2$ wander, and
$1$ never picks up the target object. No trajectory shows format collapse.
At the step level, SFT repeats its immediately preceding action on $66.7\%$
of steps (RL $43.4\%$), revisits an earlier state on $86.3\%$ ($75.2\%$),
and leaves the state unchanged on $76.2\%$ ($47.2\%$); the median number of
distinct actions per trajectory is $4$ for SFT and $8$ for RL. Trajectory-level
bootstrap intervals from $100{,}000$ resamples put the RL-minus-SFT difference
in loop share at $-12.4$ points with CI $[-28.6,1.0]$. The audit is
descriptive and makes no mechanism claim.

\paragraph{Non-arrival retention diagnostic.}
We compare two estimands on the accepted 480-row \texttt{valid\_unseen}
matrix. The endpoint estimand retains non-arrivals as zero outcomes. The
survivor-only estimand computes the paired RL-minus-SFT solver effect
separately on the states reached by each reacher, then differences those two
effects. Both use 100,000 task-stratified paired-bootstrap draws with all four
seeds and checkpoint assignments kept together. Fifty-five draws have no SFT
arrival and are excluded under the fixed empty-denominator rule, a rate of
$0.055\%$ against the $1\%$ stopping threshold.

The endpoint interaction is $24.2$ points with 95\% CI $[13.3,35.0]$. The
survivor-only estimate is $-12.6$ points with CI $[-45.8,22.7]$, and its paired
difference from the endpoint interaction is $-36.8$ points with CI
$[-68.6,-7.3]$. SFT reaches $\mathcal F_2$ in only 16 task and seed trajectories
from six tasks and has no arrivals in two of the six task types; RL reaches it
in 99 trajectories from 28 tasks. This support loss explains the wide
survivor-only interval. The diagnostic was specified after its point estimates
were observed, so it establishes neither a confirmatory effect nor a negative
conditional interaction. It shows that deleting non-arrivals does not preserve
the endpoint estimand in this setting.

\section{A Handoff State in Detail}
\label{app:case-study}

Table~\ref{tab:case-study} shows one of the $45$ \texttt{valid\_unseen}
states that the RL solver completes and the SFT solver does not, selected by a
prespecified rule (SFT-produced state, simplest task type, SFT first action
not a bridge action). The task is ``put a pencil in shelf''. The SFT reacher
produced the state by taking the pencil from the desk; the cloned observation
reads ``You pick up the pencil 1 from the desk 1'', and the solvability check
lists six bridge actions, walking to any of shelves 1 to 6, each followed by the
single winning action of moving the pencil there. Both solvers receive the
identical clone; the two handoff prompts hash to the same value. The SFT
solver's reasoning names the right plan, ``head straight to the nearest
shelf'', and then spends its first action on \texttt{look}; its second action
walks to a shelf, one action too late. The RL solver walks to a shelf and
places the pencil.

\begin{table}[h]
  \centering
  \footnotesize
  \caption{\textbf{Two solvers on one cloned state.} Bridge actions in blue.
  The SFT solver's first-step reasoning is quoted from its archived reply.}
  \label{tab:case-study}
  \begin{tabular}{@{}l p{0.42\linewidth} p{0.34\linewidth}@{}}
  \toprule
  Step & SFT solver & RL solver \\
  \midrule
  1 & \texttt{look} \newline {\itshape ``I should head straight to the nearest
  shelf. I need to locate a shelf first \ldots\ I'll scan the room.''}
    & \textcolor{blue}{\texttt{go to shelf 1}} \\
  2 & \texttt{go to shelf 1} & \texttt{move pencil 1 to shelf 1} \\
  \midrule
  Outcome & budget exhausted, not won & won \\
  \bottomrule
  \end{tabular}
\end{table}

\end{document}

%% file: tables/table1_endpoint_factorial.tex
% Table 1 -- main table.
% W = reacher (early) checkpoint, R = solver (continuation) checkpoint. Brackets give 95\% bootstrap CIs
% (B = 100{,}000). Bootstrap units differ by benchmark: TravelPlanner is
% task-paired with one seed per task; ALFWorld is stratified within six task
% types with 30 tasks x 4 seeds. Interval widths are therefore NOT comparable
% across benchmarks. Wins are not printed; wins = success_percent x n/100, with n
% the column of the same row (raw counts in data/endpoint_factorial_cells.csv).
% The TravelPlanner 7B interaction (+10.6 [0.6, 20.6]) has a
% lower bound close to zero and is the weakest of the five conditions. The
% TravelPlanner sequence 17.8 -> 14.4 -> 10.6 is monotone in scale, but the
% three intervals overlap heavily; this is an observation, not a significant trend.
\begin{tabular*}{\linewidth}{@{\extracolsep{\fill}}lr rrrr c@{}}
\toprule
& & \multicolumn{4}{c}{Endpoint success (\%)} & \\
\cmidrule(lr){3-6}
Setting & $n$ & $W$=SFT & $W$=SFT & $W$=RL & $W$=RL & Interaction \\
& & $R$=SFT & $R$=RL & $R$=SFT & $R$=RL & (pp) \\
\midrule
\multicolumn{7}{@{}l}{\emph{TravelPlanner / Agent-STAR}} \\
\quad 1.5B & 180 & 6.1 & 14.4$_{+8.3}$ & 7.8 & \textbf{33.9$_{+26.1}$} & $+$17.8 {\footnotesize[9.4, 26.1]} \\
\quad 3B & 180 & 12.2 & 24.4$_{+12.2}$ & 22.2 & \textbf{48.9$_{+26.7}$} & $+$14.4 {\footnotesize[6.1, 22.8]} \\
\quad 7B & 180 & 20.0 & 29.4$_{+9.4}$ & 38.9 & \textbf{58.9$_{+20.0}$} & $+$10.6 {\footnotesize[0.6, 20.6]} \\
\addlinespace[3pt]
\multicolumn{7}{@{}l}{\emph{ALFWorld / SkillRL}} \\
\quad Seen split & 120 & 5.0 & 9.2$_{+4.2}$ & 43.3 & \textbf{65.0$_{+21.7}$} & $+$17.5 {\footnotesize[6.7, 27.5]} \\
\quad Unseen split & 120 & 5.8 & 12.5$_{+6.7}$ & 48.3 & \textbf{79.2$_{+30.8}$} & $+$24.2 {\footnotesize[13.3, 35.0]} \\
\bottomrule
\end{tabular*}

%% file: tables/table2_reach_solve.tex
% Table 2.
% One wide table: each row is a reacher; column group (a) is its arrival rate
% and mean steps over 120 trajectories (n stated in the caption), column group
% (b) is both solvers' completion on the states that reacher produced. The
% pooled row has no (a) cells. Group (b) reports SOLVE by the policy that produced
% the state; the RL advantage appears on BOTH SFT-produced and RL-produced
% states. Row counts add up to the pooled panels (19 + 98 = 117 for the seen
% split, 16 + 99 = 115 for the unseen split). Dashes mark strata for which no
% bootstrap CI was computed (n too small after stratification); the pooled CI
% is not substituted for them. Split names follow the main text ("seen split" /
% "unseen split"); the release directories valid_seen / valid_unseen are named
% only in the appendix.
\setlength{\tabcolsep}{4.5pt}
\begin{tabular*}{\linewidth}{@{\extracolsep{\fill}}ll rr rrrl@{}}
\toprule
& & \multicolumn{2}{c}{(a) REACH} & \multicolumn{4}{c}{(b) SOLVE from identical replayed states} \\
\cmidrule(lr){3-4}\cmidrule(lr){5-8}
Split & Reacher & Arrival \% & Steps & States $n$ & $R$=SFT \% & $R$=RL \% & RL $-$ SFT (pp) \\
\midrule
Seen split & SFT & 15.8 & 41.6 & 19 & 31.6 & 57.9 & 26.3 {\scriptsize---} \\
 & RL & 81.7 & 15.7 & 98 & 53.1 & 79.6 & 26.5 {\scriptsize---} \\
 & \textbf{Pooled} &  &  & 117 & 49.6 & 76.1 & 26.5 {\scriptsize[11.2, 41.3]} \\
\addlinespace[3pt]
Unseen split & SFT & 13.3 & 44.0 & 16 & 43.8 & 93.8 & 50.0 {\scriptsize[15.4, 83.3]} \\
 & RL & 82.5 & 17.8 & 99 & 58.6 & 96.0 & 37.4 {\scriptsize[27.7, 47.0]} \\
 & \textbf{Pooled} &  &  & 115 & 56.5 & 95.7 & 39.1 {\scriptsize[29.8, 48.4]} \\
\bottomrule
\end{tabular*}

%% file: tables/table5_action_stage.tex
% Generated by scripts/make_action_stage_table.py
\begin{tabular*}{\linewidth}{@{\extracolsep{\fill}}ll rrr rr@{}}
\toprule
Split & Solver & Bridge \% & Finish $\mid$ bridge \% & Complete \% & RL only & SFT only \\
\midrule
Seen split & SFT & 72.6 {\scriptsize[64.1, 80.3]} & 68.2 {\scriptsize[57.6, 77.6]} & 49.6 {\scriptsize[40.2, 59.0]} & 41 & 10 \\
 & \textbf{RL} & 92.3 {\scriptsize[87.2, 96.6]} & 82.4 {\scriptsize[75.0, 88.9]} & 76.1 {\scriptsize[68.4, 83.8]} & & \\
\addlinespace[3pt]
Unseen split & SFT & 72.2 {\scriptsize[63.5, 80.0]} & 78.3 {\scriptsize[68.7, 86.7]} & 56.5 {\scriptsize[47.8, 65.2]} & 45 & 0 \\
 & \textbf{RL} & 97.4 {\scriptsize[93.9, 100.0]} & 98.2 {\scriptsize[95.5, 100.0]} & 95.7 {\scriptsize[91.3, 99.1]} & & \\
\bottomrule
\end{tabular*}

%% file: tables/table4_recomposition.tex
% Table 4.
% (i) The overall arrival row (13.3 / 82.5) matches alfworld_reach.csv on
%     valid_unseen exactly: REACH is measured on the target split.
% (ii) The SOLVE panel is collected independently (n = 20 per task type per
%      reader) and does not overlap the 115 replayed states of Table 2. This
%      non-overlap is what makes the two measurements independent.
% (iii) The overall row is the UNWEIGHTED MEAN over the six task types
%       (124/6 = 20.67 predicted, 145/6 = 24.17 observed), not a pooled
%       recomputation. This matches the stratified bootstrap unit
%       (task_stratified_within_six_task_types); readers who pool instead will
%       not reproduce these numbers.
% Dashes mark cells the source CSV leaves empty for the overall row.
\begin{tabular}{l rr rr rrr}
\toprule
& \multicolumn{2}{c}{Target-split arrival \%}& \multicolumn{2}{c}{External SOLVE \%}& \multicolumn{3}{c}{Interaction (pp)} \\
\cmidrule(lr){2-3}\cmidrule(lr){4-5}\cmidrule(lr){6-8}
Task type & SFT & RL & SFT & RL & Pred. & Obs. & Resid. \\
\midrule
look at obj in light & 20.0 & 80.0 & 10.0 & 35.0 & 15.0 & 40.0 & 25.0 \\
pick \& place & 35.0 & 70.0 & 85.0 & 100.0 & 5.2 & $-$5.0 & $-$10.2 \\
clean then place & 10.0 & 70.0 & 20.0 & 100.0 & 48.0 & 60.0 & 12.0 \\
cool then place & 0.0 & 100.0 & 80.0 & 100.0 & 20.0 & 25.0 & 5.0 \\
heat then place & 15.0 & 100.0 & 60.0 & 80.0 & 17.0 & 5.0 & $-$12.0 \\
pick two \& place & 0.0 & 75.0 & 60.0 & 85.0 & 18.8 & 20.0 & 1.2 \\
\midrule
\textbf{Overall} & 13.3 & 82.5 & --- & --- & 20.7 & 24.2 & 3.5 \\
\bottomrule
\end{tabular}

%% file: tables/table3_budget.tex
% Table 3.
% Budget data exist for valid_seen ONLY. Source: protocol
% skillrl-alfworld-natural-d2-frontier-budget-curve-val30-s4-v1, derived from
% skillrl-alfworld-7b-natural-d2-segment-aware-confirm-validseen30-s4-v2.
% No accepted valid_unseen budget artefact exists locally, so these intervals
% must not be attached to the valid_unseen arrival curve of Figure 2 (left).
% The bolded row is a cross-budget counterfactual (RL with 8 actions vs SFT with
% 48), not a same-budget comparison: it shows the REACH gap is not an artefact
% of action budget. B = 100{,}000 bootstrap resamples.
% Claim boundary (from the source artefact): cumulative frontier arrival within
% at most 48 writer actions under frozen policies; no extrapolation to unbounded
% compute, additional sampling, search trees, or explicit short-deadline
% replanning.
\begin{tabular}{lrrr}
\toprule
\multicolumn{4}{l}{\emph{(a) Arrival by reacher action budget, seen split}} \\
\addlinespace[1pt]
Budget (steps) & Reacher & Arrival \% & 95\% CI \\
\midrule
8 & SFT & 10.8 & [4.2, 17.5] \\
8 & RL & 47.5 & [34.2, 60.0] \\
16 & SFT & 14.2 & [7.5, 20.8] \\
16 & RL & 75.8 & [62.5, 88.3] \\
48 & SFT & 15.8 & [8.3, 22.5] \\
48 & RL & 81.7 & [69.2, 93.3] \\
\addlinespace[3pt]
\midrule
\multicolumn{4}{l}{\emph{(b) Contrasts}} \\
\addlinespace[1pt]
\multicolumn{2}{l}{Comparison} & Effect (pp) & 95\% CI \\
\midrule
\multicolumn{2}{l}{RL@8 $-$ SFT@8} & 36.7 & [23.3, 50.0] \\
\multicolumn{2}{l}{RL@16 $-$ SFT@16} & 61.7 & [48.3, 74.2] \\
\multicolumn{2}{l}{RL@48 $-$ SFT@48} & 65.8 & [54.2, 77.5] \\
\multicolumn{2}{l}{\textbf{RL@8 $-$ SFT@48}} & \textbf{31.7} & [19.2, 44.2] \\
\bottomrule
\end{tabular}